\documentclass[sigconf]{acmart}
\renewcommand\footnotetextcopyrightpermission[1]{} 
\AtBeginDocument{%
  \providecommand\BibTeX{{%
    \normalfont B\kern-0.5em{\scshape i\kern-0.25em b}\kern-0.8em\TeX}}}

 \setcopyright{none} 
\begin{document}

\title{Compositional Generalization for Natural Language Interfaces to Web APIs}

\author{Saghar Hosseini}
\email{sahoss@microsoft.com}
\affiliation{%
  \institution{Microsoft Research}
  \city{Redmond}
  \state{WA}
  \country{USA}
}

\author{Ahmed Hassan Awadallah}
\email{hassanam@microsoft.com}
\affiliation{%
  \institution{Microsoft Research}
  \city{Redmond}
  \state{WA}
  \country{USA}
}

\author{Yu Su}
\email{yusu2@microsoft.com}
\affiliation{%
  \institution{Microsoft}
  \city{Redmond}
  \state{WA}
  \country{USA}
}

\begin{abstract}
This paper presents Okapi, a new dataset for Natural Language to executable web Application Programming Interfaces (NL2API). This dataset is in English and contains 22,508 questions and 9,019 unique API calls, covering three domains. We define new compositional generalization tasks for NL2API which explore the models' ability to extrapolate from simple API calls in the training set to new and more complex API calls in the inference phase. Also, the models are required to generate API calls that execute correctly as opposed to the existing approaches which evaluate queries with placeholder values. Our dataset is different than most of the existing compositional semantic parsing datasets because it is a non-synthetic dataset studying the compositional generalization in a low-resource setting. Okapi is a step towards creating realistic datasets and benchmarks for studying compositional generalization alongside the existing datasets and tasks. We report the generalization capabilities of sequence-to-sequence baseline models trained on a variety of the SCAN and Okapi datasets tasks. The best model achieves 15\% exact match accuracy when generalizing from simple API calls to more complex API calls. This highlights some challenges for future research. Okapi dataset and tasks are publicly available at \it{https://aka.ms/nl2api/data}. 
\end{abstract}



\begin{teaserfigure}
\centering
  \includegraphics[width=0.85\textwidth]{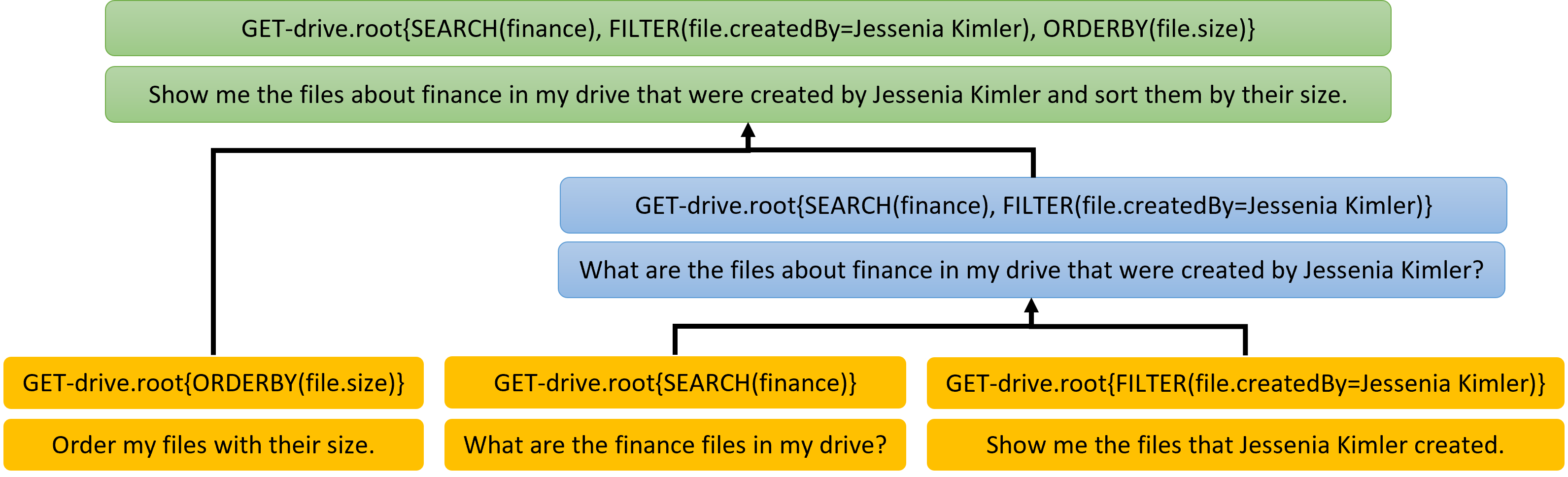}
  \caption{\small{The compositional nature of web APIs. Each level consists of abstract representations of API queries (top) and their corresponding user's natural language commands (bottom). As we move from bottom to the top of this graph, we can create new API calls from different combination of the lower levels.}}
  \label{fig:teaser}
\end{teaserfigure}

\maketitle
\section{Introduction}
\label{sec:intro}
Web APIs are application programming interfaces over the Web which can be accessed via HTTP protocol. Web API is often used to provide an interface for Web sites and can be used to access data from a database and save data back to the database. These interfaces have become ubiquitous by enabling open platforms that any developer can interact with. Currently, there are about 24,000 Web APIs available\footnote{https://www.programmableWeb.com/apis/directory} for a wide range of domains including productivity, social media, payment, shopping, education, messaging, science, games, and maps.  As the number of available Web APIs rapidly increases, the adaptation cost on users also increases. This opens an interesting avenue for human-machine interface for Web APIs. There has been some efforts in leveraging natural language for generating API calls \citep{Su2017,Su2018}, designing and building  RESTful API services~\citep{Dimanidis2018}, and Web API service recommendation~\citep{Lin2018NL2APIAF}. Recently, \citet{Degbelo2020OpenGR} performed a user study on the learnability of order-agnostic and natural-language based API design. They collected data from 20 participants for geo-referenced open datasets and their study showed that Natural Language Interface (NLI) approach on top of RESTful APIs is easily learnable and it can reduce users' effort on learning new APIs.  
Therefore, in this work, we focus on Natural Language to executable Web Application Programming Interfaces (NL2API)~\citep{Su2017} which can understand the meaning of natural language sentences and map them to meaningful executable API queries. Such systems have a great impact on empowering users by making it effortless to leverage these open platforms.

Web APIs support optional query parameters that can be used to specify and control the returned data in a response as shown in Figure~\ref{fig:teaser}. There is usually a large number of query parameters in individual Web APIs and seldom enough training data to cover all the possible combinations of these parameters. Moreover, there are infinite ways of expressing an API call in natural language. Therefore, NL2APIs are facing the low-data settings challenge. In order to address this issue and build low-cost NL2APIs, these systems must have the ability to generalize to out-of-distribution API calls after being exposed to a limited set of examples. For example, the training data includes “A”, “BC”, and the model is expected to generalize to “ABC” in inference time. This ability is called compositional generalization. 

Recently, there has been a growing interest in compositional generalization in the NLIs to other formal representations~\citep{Lake2018GeneralizationWS, Keysers2020MeasuringCG,liu2020compositional,nye2020,chen2020compositional,Herzig2020SpanbasedSP,Shaw2020CompositionalGA} which enables these systems to systematically compose complex examples after being exposed to simple components during training. This is an ability demonstrated by human intelligence which is often lacking in state-of-the-art neural network models.

Due to the lack of realistic benchmarks that comprehensively measure compositional generalization, we are facing a challenge in evaluating NL2APIs with respect to this ability . \citet{Shaw2020CompositionalGA} highlighted in their work that evaluating NLIs on a diverse set of benchmarks is important and the performance on \textit{synthetic} datasets~\citep{Lake2018GeneralizationWS, Keysers2020MeasuringCG} is not well-correlated with performance on non-synthetic tasks with natural language variation. Synthetic datasets, in which both natural language questions and queries were generated by a program, are very useful in zero-shot setting. However they do not fully represent the variations in natural language. Moreover, most of the existing approaches ignore the importance of generating executable queries, which is crucial for creating an end-to-end NLIs. In their evaluation, the entities or values are replaced by placeholders or generic terms. 

Therefore, to address the need for a diverse and realistic dataset to study the compositional generalization in an end-to-end NL2API, we introduce Okapi and two compositional generalization tasks in Web APIs. To measure the NL2APIs ability to generalize to more complex API calls, we split Okapi to short or simple API calls for training set and long or complex API calls for inference time in first task. For example, the training set contains "A", "B", "C", "D", and their simple combinations "AB" and "CD". In this task, NL2API must generalize to "ABCD" at inference time. In second task, we measure the NL2API ability in generalizing to unseen combinations of API parameters. For example, NL2API must generalize to "BC" and "AD" at inference time. 

In summary our key contributions are as follows: (1) we created a human-labeled large-scale dataset for building NL2APIs. The API calls were generated based on real world scenarios in Microsoft Graph API and in three different domains. (2) we Defined two compositional generalization tasks on our dataset which require the ability to generalize to more complex API calls and to unseen combination of parameters in API calls. (3) To evaluate these tasks difficulty, we experiment with several general-purpose sequence-to-sequence models which are frequently used for NLIs and perform competitively with the publicly available state of the art models. These models were evaluated based on their ability to generate the correct \textit{executable} API queries.

\section{Related Work and Existing Datasets}\label{sec:related_work}
Semantic parsing techniques have been developed for several formal languages such as logical forms \citep{price-1990-evaluation, Zelle1996}, SQL~\citep{Yu&al.18c, zhongSeq2SQL2017}, SPARQL~\citep{Keysers2020MeasuringCG, berant-etal-2013-semantic}, Web API~\citep{Su2017,Su2018}, and dataflow programs~\citep{SMDataflow2020} for building NLIs to knowledge bases, accessing resources, question answering and instructions following in smart agents, code generation, and many other applications. NL2API is also a semantic parsing task that maps natural language utterances to executable Web APIs. In the following section, we review the existing benchmarks and methods that study compositional generalization in different NLIs.

\subsection{Datasets and Benchmarks}
\par Several synthetic datasets and tasks have been created for investigating compositional generalization in NLIs including SCAN~\citep{Lake2018GeneralizationWS} which has become the primary benchmark in this field. SCAN is a corpus of natural language commands and action sequence pairs, for example, "turn left and jump" maps to "{\small\texttt{I-TURN-LEFT I-JUMP}}". There are several methods of splitting SCAN dataset to evaluate compositional generalization in different aspects. In this work, we focus on the performance of models on SCAN in the Length~\citep{Lake2018GeneralizationWS} and Maximum Compound Divergence (MCD)~\citep{Keysers2020MeasuringCG} splits. The Length split separates the training and test sets such that all the examples in test set have longer action sequence. MCD was proposed by ~\citet{Keysers2020MeasuringCG} along with CFQ, a synthetic natural language and SPARQL pair dataset. This split is based on the distribution of individual rules (atoms) and their combinations (compounds) in examples. MCD splits the dataset to train and test sets such that the divergence between the distribution of compounds in train and test sets is maximized while the atoms distribution divergence remains small.  

Moreover, another synthetic dataset called COGS~\citep{kim2020cogs} has been developed that maps sentences to logical forms and evaluates the structural and lexical generalizations. 
\citet{Shaw2020CompositionalGA} leveraged the GeoQuery dataset~\cite{Zelle1996} to evaluate the compositional generalization on a non-synthetic dataset. GeoQuery contains questions about US geography and their representation in Functional Query Language and was split according to Length and MCD methods. 

The task of building an end-to-end framework for transferring natural language to a given Web API was introduced by \citet{Su2017,Su2018}. They created a non-synthetic dataset for email and calendar domains via crowd-sourcing and evaluated sequence-to-sequence and language models against their dataset. However, their dataset is not publicly available and includes fewer domains and is smaller in size compare to Okapi. Also, it lacks diversity in questions and parameter values, and therefore is not a good representative of realistic API calls.   
We created Okapi with three criteria in mind: (1) being realistic and a good representative of actual applications, (2) having diversity in both API calls and natural language questions, and (3) high quality with respect to annotations. 
\subsection{Methods}
Several approaches have been developed to solve the compositional generalization tasks in the datasets mentioned in the previous section. Perfect accuracy has been achieved for SCAN splits by several specialized architecture models~\citep{Shaw2020CompositionalGA, nye2020, liu2020compositional,chen2020compositional} which learn and incorporate grammar or structure to tackle compositional generalization challenges.

\citet{Herzig2020SpanbasedSP} and \citet{Shaw2020CompositionalGA} both evaluated their models on a subset of GeoQuery dataset splits and achieved strong performance. However the span-based semantic parsing model \citep{Herzig2020SpanbasedSP} requires manual task specific pre-processing. Moreover, \citet{Oren2020ImprovingCG}, studied the compositional generalization on several non-synthetic text-to-SQL datasets, which proved to be challenging. Note that these models are evaluated on examples which use placeholders in place of their entities' values. Moreover, at this time none of these systems have publicly available code to be tested on Okapi dataset. 

Finally, \citet{Furrer2020CompositionalGI} investigated the ability of pre-trained transformer language models in compositional generalization and found them helpful while not enough to solve this problem. In addition, they evaluated several specialized architectures and discovered those models do not transfer well to other datasets.

\section{Dataset Creation and Statistics}\label{sec:dataset}
In creating Okapi, we focus on mapping natural language questions or commands to RESTful APIs. A RESTful API is a style of application program interfaces (APIs) that uses HTTP requests to access and use data. They support multiple operations, namely {\small\texttt{GET, PUT, POST}} and {\small\texttt{DELETE}} corresponding to the reading, updating, creating and deleting operations of the underlying resources.

More specifically, we adopt a set of APIs from the the Microsoft Graph API suite that enables users to interact with their emails, calendar and files (on OneDrive) and follows the Open Data Protocol (OData) ~\citep{chappell2011introducing}.  OData is a standard protocol for providing a generic way to organize and describe data based on the Entity Data Model where each resource is represented with an entity that has a set of properties. For example, the {\small\texttt{event}} entity, representing a calendar event has properties like organizer, attendees, start time, end time, etc. The protocol also defines a set of parameterized query options designed to enable retrieval and manipulation of the data such as {\small\texttt{FILTER}, \texttt{SEARCH}, \texttt{SELECT}, \texttt{ORDERBY}}, etc. A full list of query options is shown in Table~\ref{tab:API_properties}.
\begin{table*}
\centering
\begin{tabular}{@{}ll@{}}
\toprule
\centering
{\small\textbf{Parameters}} & {\small \textbf{Description}} \\ \midrule
{\small\texttt{SEARCH(String)}}   & {\small Search for entities containing specific keywords}\\
{\small\texttt{FILTER(BoolExpr)}} & {\small Filter resources by some criteria}  \\
{\small\texttt{ORDERBY(Property)}}  & {\small Sort the property in ’asc’ or ’desc’ order}   \\
{\small\texttt{SELECT(Property)}} & {\small Return a certain property} \\
{\small\texttt{COUNT()}} & {\small Count the number of matched resources} \\
{\small\texttt{TOP(Integer)}} & {\small Only return the first certain number of results} \\
\bottomrule
\end{tabular}
\caption{\label{tab:API_properties}  \small List of API query options used in our dataset and defined by the OData protocol}
\end{table*}

Therefore, an API call consists of an HTTP request (e.g. {\small \texttt{GET}}), an end point (e.g. \textit{https://graph.microsoft.com}), the resource being requested (e.g. files) and a list of parameterized query options (e.g. {\small \texttt{SELECT(file.name)}, \texttt{FILTER(file.createdBy=John Smith)}}).


Note that Web APIs are highly compositional which makes them a good domain for evaluating the compositional generalization of semantic parsing models. For example, Figure~\ref{fig:teaser} shows how we can combine multiple API calls to create a more complex API call.

\subsection{Dataset Construction}
 We developed Okapi in three steps which required about 700 hours of annotation and reviewing. First we constructed the API calls, then we asked annotators to provide natural questions for those API calls. Lastly, the crowd workers generated paraphrases for the utterances collected in the first step. This was to increase the diversity. We elaborate on each of these three steps in the following sections.
 \\
\subsubsection{{Generating API Calls}}~\\
The API calls were generated following the same approach as in~\citep{Su2017} for three domains including documents, calendar, and emails. In text to SQL and SPARQL tasks, significant improvements in the performance were achieved using intermediate representation \citep{Furrer2020CompositionalGI, guo-etal-2019-towards}. Similarly, we used an intermediate representation of the RESTful API (API call in Figure~\ref{fig:teaser}). This intermediate representation can be converted to real API calls in a deterministic way and removes the unnecessary information such as URL format, HTTP headers, etc. from the formal representation. We refer to this intermediate representation as API call in the remainder of this paper. 

API calls were generated based on the specification of an API. We enumerated all the combinations of query options, properties, and property values to generate API calls. Moreover, simple rules can be used to make sure the combination of query options are reasonable. For example, {\small\texttt{TOP}} has to be applied on a sorted data and thus {\small\texttt{ORDERBY}} is required. In addition to the query options, we need property values to generate realistic API calls. To populate the property values, we leveraged the public datasets for names\footnote{https://github.com/smashew/NameDatabases}, key phrases\footnote{https://github.com/LIAAD/KeywordExtractor-Datasets}, and email categories\footnote{https://catalog.ldc.upenn.edu/LDC2015T03}. For example, in our dataset, the {\small\texttt{SEARCH}} and {\small\texttt{FILTER}}  parameters are associated with specific values, e.g. {\small\texttt{SEARCH(finance)}} instead of {\small\texttt{SEARCH(placeholder)}}. 
\\
\subsubsection{{Natural Utterance Annotation}}
\begin{figure}[hb]
    \centering
    \includegraphics[width=0.99\linewidth]{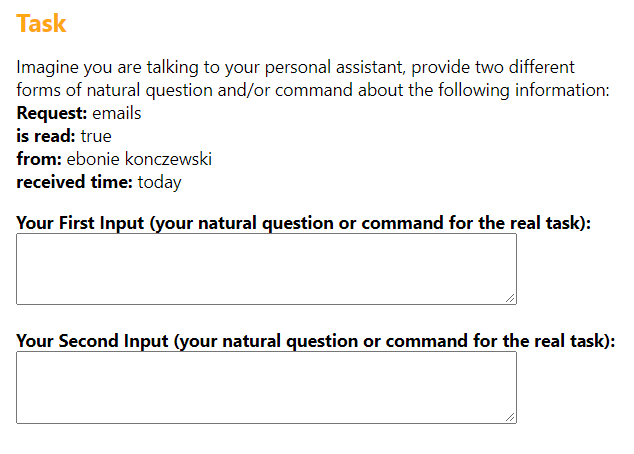}
    \vspace{-0.7cm}
    \caption{\small A screenshot of the annotation interface with an example annotation task. }
    \label{fig:uhrs}
    \vspace{-0.3cm}
\end{figure}

~\\
We used a crowd-sourcing platform to hire a skilled group of annotators who were proficient in English. The dataset was created by 20 annotators and they were paid \$13/hour to encourage them to focus on the quality of the annotation instead of maximizing their hourly pay. We trained the group of annotators using guidelines and training examples to be prepared for this task. Moreover, they were allowed to spend three minutes on each example. For each example we asked one annotator to provide two natural questions or commands that represent the information provided by the API call to improve the diversity in natural utterances. An example of an annotation task is shown in Figure~\ref{fig:uhrs}. Moreover, we used a template based method to explain the API calls in a natural way. The dataset was collected in small batches ranging from 100 to 1000 and after each batch was completed a subset of the annotations were manually reviewed by our team. We checked if the question or command reflects the meaning of its corresponding API call. Then we removed the low quality annotations and their respective annotators from the task and sent their annotations to be re-judged. 
 
\subsubsection{Paraphrase Natural Utterance }
 After reviewing the dataset we asked a different set of annotators to generate paraphrases for some of the questions or commands to improve the diversity of our dataset. We collected the paraphrases in small batches ranging from 100 to 500 and followed the same quality control policy as in previous step. 
\subsection{Dataset Summary and Comparison}\label{sec:dataset_summary}

\begin{table*}[ht]
\centering
\begin{tabular}{@{}lcccccc@{}}
\toprule
\centering
 {Dataset} &  {\#Questions} &{ \#Queries} & {\#Domains} & {\#Templates} & { Realistic} & { 2-grams Jaccard Similarity}\\ \midrule
{CFQ} & { 239,357} & { 228,149}  & - & { 34921} & { No} & { 0.04} \\
{SCAN}  & { 20,910} & { 20,910} & { 1} & -  & { No} & { 0.39} \\
{GeoQuery} & { 880} & { 246} & { 1} & { 98} & { Yes} & { 0.24}  \\
{\textbf{Okapi}} & { 22628} & { 9019} & { 3} & { 1961} & { Yes} & { 0.14} \\
\bottomrule
\end{tabular}
\caption{\label{tab:data_comparison} \small Okapi complexity statistics in comparison to other semantic parsing datasets  }
\end{table*}

\begin{figure*}[ht]
    \centering
    \includegraphics[width=\textwidth]{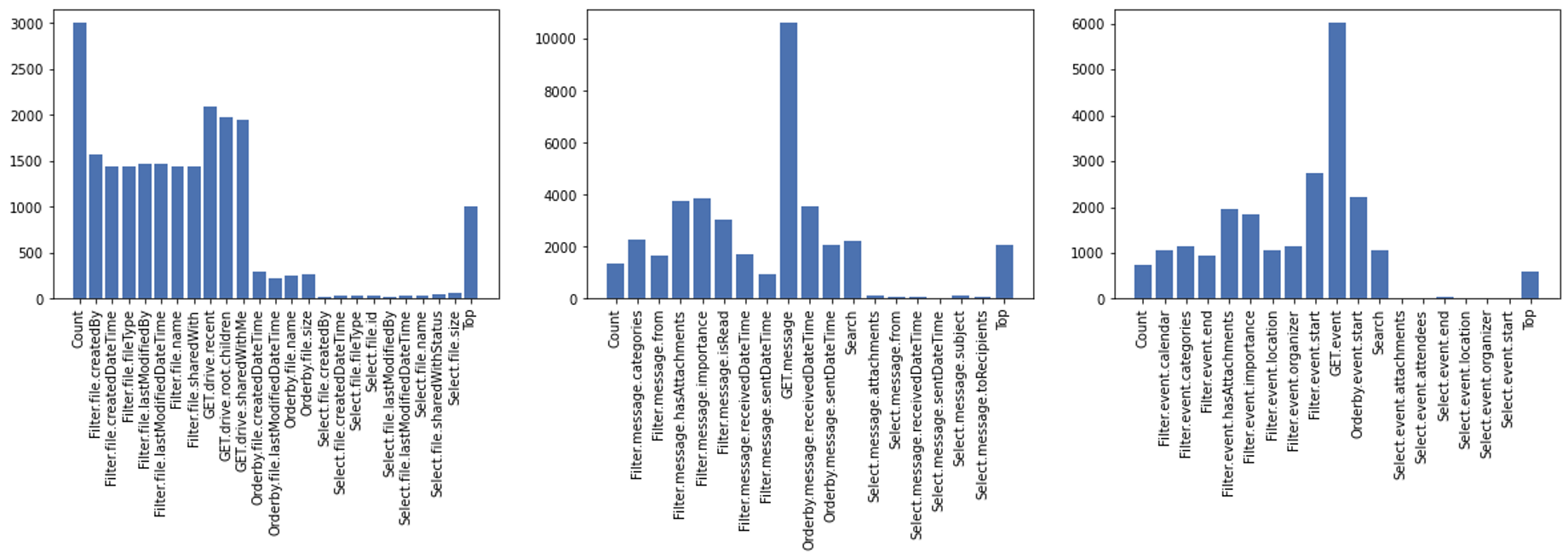}
    \caption{\small Distribution of API parameters and their properties in document, email, and calendar (left to right).}
    \label{fig:parameters_dist}
\end{figure*}

We summarized the distribution of the Okapi dataset for different domains in Table~\ref{data_dist}. Number of parameters refers to the parameters defined in Table~\ref{tab:API_properties}. In addition, the distribution of API parameters and their properties are presented in Figure~\ref{fig:parameters_dist}. Overall, we have 5885 natural language API call pairs in the document domain out of which, 3005 are unique API calls. The email domain has 10612 natural language API call pairs where 3007 API calls are unique and calendar domain has 6011 natural language API call pairs and 3007 unique API calls. We sampled 500 examples from each domain and examined the annotation. The annotation error in all the domains is about 3\%. Note that the annotation errors includes mistakes regarding spelling, missing information and providing wrong information. 
\begin{table}[ht]
\centering
\begin{tabular}{@{}lcccccc@{}}
\toprule
\centering
{\small\# of Parameters} & {\small 2} & {\small 3} & {\small 4} & {\small5} & {\small6} & {\small7 }\\ \midrule
{\small Document}  & {\small 751} & {\small 1697} & {\small 1963} & {\small 1104} & {\small 336} & {\small34}  \\
{\small Email} &  {\small 412} & {\small 2301} & {\small 7793} & {\small 106} & {\small 0} & {\small 0}   \\
{\small Calendar}  & {\small 186} & {\small 1144} & {\small 4621} & {\small60} & {\small0} & {\small 0}   \\
\bottomrule
\end{tabular}
\caption{\label{data_dist} \small API call distributions in Okapi corpus. }
\end{table}
In our data annotation, we first ask crowd workers to annotate a canonical utterance for each API call and then ask a different set of crowd workers to paraphrase the canonical utterance in order to improve diversity. A potential drawback of such an annotation scheme, if not controlled carefully, is that the paraphrases may be biased towards the canonical utterance \citep{herzig-berant-2019-dont} and they end up being too similar and less diverse. 
This is quite common in crowd-powered paraphrasing. However, that may not be the case for Okapi. We show that in Okapi, the phrases and their corresponding canonical utterance are quite dissimilar in wording. We measured the average Jaccard similarity and Levenshtein edit distance between each natural utterance provided by human and corresponding canonical utterance obtained from API call (all lower-cased). The average Jaccard similarity is 0.19 and 0.06 for uni-grams and bi-grams. The average Levenshtein edit distance is 60.5 with average length of 96 and 83.2 characters for canonical utterances and questions, respectively. We think this is a decent level of linguistic diversity since a 0.19 Jaccard similarity of uni-grams means that on average only around 19\% of exact words are shared between each pair of canonical utterance and paraphrase.

Moreover, we summarize the statistics of Okapi and several other semantic parsing datasets in Table~\ref{tab:data_comparison}. The user utterances in Okapi are collected from human annotators while SCAN and CFQ user utterances are program generated (synthetic).  Even though Okapi dataset is much smaller than the synthetic datasets, it is larger than GeoQuery. 

In addition, Okapi is more diverse with respect to the domains and API call templates. To define the templates in Okapi dataset, we ignored the API property values and only considered the combinations of parameters and properties. For example, the template for the API call in Figure~\ref{fig:teaser} is a set constructed of {\small\texttt{SEARCH(placeholder)}}, {\small\texttt{ORDERBY(placeholder)}}, and {\small\texttt{FILTER(file.CreatedBy=placeholder)}}. 

Note that the templates are order-invariant with respect to API parameters.  Moreover, the 2-grams Jaccard similarity in Table~\ref{tab:data_comparison} is measured for queries with the same template to represent the linguistic diversity in each dataset. 

\section{The NL2API Tasks}\label{sec:nl2api_tasks}
In addition to the proposed dataset, we define two NL2API task setups. These tasks explore different aspects of compositional generalization in NL2API similar to previous work~\citep{Lake2018GeneralizationWS, finegan-dollak-etal-2018-improving, Oren2020ImprovingCG}. In our tasks, we evaluate the models performance on generating values and predicting the correct executable API. 

\subsection{Generalize to more Complex APIs }
\begin{figure}[h]
    \centering
    \includegraphics[width=0.85\linewidth]{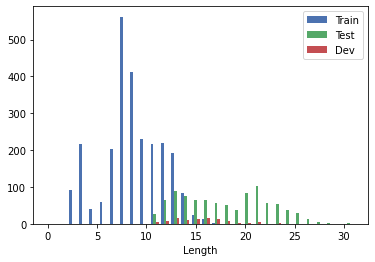}
    \caption{\small Document API call distribution in \emph{Length} split. }
    \label{fig:doc_vertical_distribution}
\end{figure}

\begin{figure}[h]
    \centering
    \includegraphics[width=0.85\linewidth]{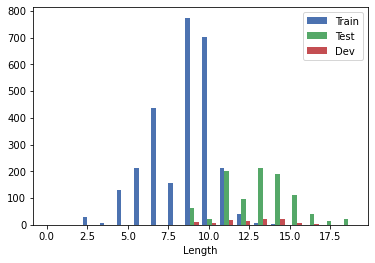}
    \caption{\small Email API call distribution in \emph{Length} split. }
    \label{fig:emai_vertical_distribution}
\end{figure}

\begin{figure}[h]
    \centering
    \includegraphics[width=0.85\linewidth]{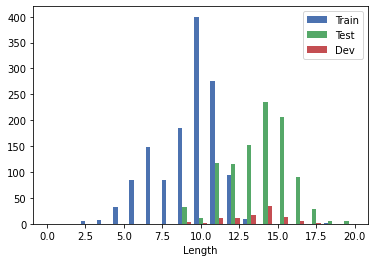}
    \caption{\small Calendar API call distribution in \emph{Length} split. }
    \label{fig:event_vertical_distribution}
\end{figure}

\begin{figure}[h]
    \centering
    \includegraphics[width=0.85\linewidth]{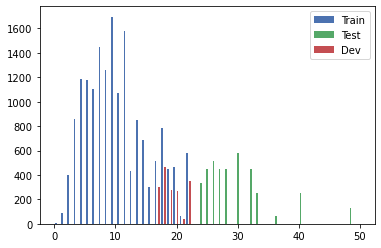}
    \caption{\small SCAN target domain length distribution in \emph{Length} split.}
    \label{fig:scan_vertical_distribution}
\end{figure}

In this task, we study the systematic form of generalization, in which the models must extrapolate to APIs more complex than the ones the models have already seen. We focus on the number of parameters in the API call where the training set is limited to two and three parameters APIs. Then we randomly select 100 examples from the four or five parameters APIs for the development set and about 1000 examples with four to seven API parameters for the test set. The distribution of number of parameters in train, development, and test sets are available in Table~\ref{tab:vertical_dist}. Each value in the table, represents the number of natural sentence and API pairs. We assumed that it is reasonable for the user to generate a small set for validation with longer API calls and hence we have an overlap between development and test set with respect to the number of parameters. Moreover, we made sure all the API parameters appear in training set. 
\begin{table}
\centering
\begin{tabular}{@{}l|cc|cc|cccc@{}} 
\toprule
\centering
& \multicolumn{2}{c|}{{\small \textbf{Train}}} & \multicolumn{2}{c|}{{\small\textbf{Dev}}} & \multicolumn{4}{c}{{\small\textbf{Test}}}\\ 
{\small \# Parameters} & {\small 2} & {\small 3} & {\small 4} & {\small 5} & {\small 4} & {\small 5} & {\small 6} & {\small 7} \\ \midrule
{\small Document}  & {\small 751} & {\small 1697} & {\small 64} & {\small 36} & {\small 372} & {\small 170} & {\small 336} & {\small 34}  \\
{\small Email} &  {\small 412} & {\small 2301} & {\small 100} & {\small 0} & {\small 866} & {\small 106}   & {\small 0} & {\small 0}   \\
{\small Calendar}  & {\small 186} & {\small 1144} & {\small 100} & {\small 0} & {\small 940} & {\small 60}  & {\small 0}  & {\small 0}    \\
\bottomrule
\end{tabular}
\caption{\label{tab:vertical_dist}\small API call distributions in \emph{Length} split. }
\end{table}
It is worth noting that this task correlates with the \emph{Length} split in SCAN dataset~\citep{Lake2018GeneralizationWS} in which the train-test split is based on the number of actions in each command. Therefore we refer to this task as  \emph{Length} split. However, the \emph{Length} split in Okapi is based on the number of API parameters and is different than the \emph{Length} split in SCAN which is based on the number of tokens in the queries. For comparison we analyzed the length distribution of API calls in train, development, and test sets and observed an overlap between them (Figures~\ref{fig:doc_vertical_distribution} to \ref{fig:scan_vertical_distribution}). Note that the length is the number of words in the API call and this overlap is due to the longer parameter values in some of the examples in train set. Another observation is that the target sequences in Okapi dataset are generally shorter than SCAN dataset with the maximum length of 33 words. However, our task differs from SCAN \emph{Length} split with respect to the output vocabulary and having arbitrary values for API property values. Note that, SCAN dataset vocabulary is limited to {\small\texttt{I-JUMP, I-LOOK, I-WALK, I-TURN-RIGHT, I-TURN-LEFT}}, and {\small\texttt{I-RUN}}. 

\subsection{Generalize to Unseen API Templates}
\label{subsec:horizontal task}
Our next task is closest to the \emph{Program} split in \citet{finegan-dollak-etal-2018-improving} and \citet{Oren2020ImprovingCG}, in which the model is exposed to a finite set of API calls templates and must bootstrap to unseen templates. Therefore, we call this task \emph{Program} split. Templates in Okapi dataset are the possible combinations of parameters and properties as explained in Section~\ref{sec:dataset_summary}. We randomly split the dataset based on the template such that there is no overlap between the templates in train, validation and test sets. The distribution of templates in each domain is presented in Table~\ref{tab:horizontal_template_distl} which shows an allocation of 50\%, 25\%, and 25\% with respect to the number of templates in train, development, and test sets, respectively. The statistics of \emph{Program} split is presented in Table~\ref{tab:horizontal_datasize}. Moreover, we observe similar distributions with respect to the number of parameters across train, development, and test sets in all the domains.
\begin{table}[h]
\centering
\begin{tabular}{@{}lccc@{}} 
\toprule
\centering
{\small Set} & {\small \textbf{Train}} & {\small \textbf{Dev}} & {\small \textbf{Test}}\\ 
\midrule
{\small Document}  & {\small 750} & {\small 375} & {\small 375}  \\
{\small Email} & {\small 96} & {\small 49} & {\small 49} \\
{\small Calendar}  & {\small 135} & {\small 69} & {\small 66}\\
\bottomrule
\end{tabular}
\caption{\label{tab:horizontal_template_distl}\small Number of API templates in each set for \emph{Program} split. }
\end{table}
\begin{table}[h]
\centering
\begin{tabular}{@{}lccc@{}} 
\toprule
\centering
{\small Set} & {\small \textbf{Train}} & {\small \textbf{Dev}} & {\small \textbf{Test}}\\ 
\midrule
{\small Document}  & {\small  2934} & {\small 1362} & {\small 1589}  \\
{\small Email} & {\small  5300 } & {\small 2514} & {\small 2791} \\
{\small Calendar}  & {\small 3007 }& {\small1414} & {\small 1228} \\
\bottomrule
\end{tabular}
\caption{\label{tab:horizontal_datasize}\small Dataset sizes in \emph{Program} split. }
\vspace{-0.5cm}
\end{table}

\section{Methods}\label{methods}
\begin{table*}[h]
\centering
\begin{tabular}{@{}lcccccccc@{}} 
\toprule
\centering
& \multicolumn{2}{c}{{\small \textbf{SCAN}}} & \multicolumn{2}{c}{{\small \textbf{Okapi}-Doc}} & \multicolumn{2}{c}{{\small\textbf{Okapi}-Email}} &
\multicolumn{2}{c}{{\small\textbf{Okapi}-Calendar}}\\ 
{\small Method} & {\small Len} & {\small MCD} & {\small Len} & {\small Program} & {\small Len} & {\small Program} & { \small Len } & {\small Program}  \\
\midrule
{\small LSTM+Attention} & {\small 14.1} & {\small 6.1} & {\small 0} & {\small 35.10} & {\small 0} & {\small 26.0} & {\small 0} & {\small 34.0}\\
{\small Transformer+Copy} & {\small 0} & {\small 0} & {\small 7.14}& {\small \textbf{83.2}} & {\small 11.2} & {\small \textbf{70.5}} & {\small 10.6} & {\small \textbf{81.8}}\\
{\small T5-Base} & {\small \textbf{14.4}} & {\small \textbf{15.4}} & {\small \textbf{15}} & {\small 31.37} & {\small \textbf{14.85}} & {\small 41.06} & {\small \textbf{13.2}} & {\small 25.79}\\
\bottomrule
\end{tabular}
\caption{\label{tab:results}\small Comparisons of the SCAN and Okapi datasets with respect to exact match accuracy(\%).  }
\end{table*}
In order to asses the difficulty of the NL2API tasks defined in Section~\ref{sec:nl2api_tasks}, we experiment with several neural semantic parsing models that are competitive with publicly available systems for different semantic parsing tasks. We only consider general sequence to sequence architectures and large pre-trained language models that can be applied to a wide range of formal languages such as SQL, SPARQL and API. 
The only method with perfect scores on all SCAN split which doesn't require task specific resources is NQG-T5~\citep{Shaw2020CompositionalGA}. However, its performance is limited by the inductive grammar model which naturally perform well on synthetic datasets. 
\\
\textbf{LSTM+Attention} \quad Similar to \citet{Keysers2020MeasuringCG} we train an LSTM with attention mechanism. We also use value copying mechanism~\citep{gu-etal-2016-incorporating, jia-liang-2016-data} in the decoder due to the nature of Web API calls. At each step, the decoder generates a distribution over possible API parameters and properties and copying a token from the natural utterance.  
\\
\textbf{Transformer+Copy}\quad This is a general-purpose and frequently used model in semantic parsing field which consists of a pre-trained BERT encoder~\citep{devlin-etal-2019-bert} and an autoregressive Transformer decoder~\citep{Vaswani2017AttentionIA} that attends over the outputs of the encoder. \citet{suhr-etal-2020-exploring} showed that this model performs on par with respect to the more complex public models for Text-to-SQL tasks and specifically on Spider~\citep{Yu&al.18c} dataset. This model generates executable SQL queries without placeholder values by leveraging attention based copying mechanism \footnote{ We used the implementation of Transformer+Copy released by \citep{suhr-etal-2020-exploring}}. In our setup, we did not include the database schema in model's input. We use the validation set for hyper-parameter tuning and report those hyper-parameter values in the supplementary material section. 
\\
\textbf{T5-Base}\quad This is a pre-trained transformer language model which reframes all Natural Language Processing (NLP) tasks as a text-to-text problem~\citep{T5} and has been previously evaluated on semantic parsing datasets~\citep{Furrer2020CompositionalGI}. We finetuned T5-Base \footnote{We used the T5 implementation in Transformers library of Hugging Face (https://huggingface.co/transformers)} with a learning rate of $3e^{-4}$, batch size of 32, and up to 200k steps. However, we choose the number of steps based on the performance of the model on validation set.

\section{Experiments and Discussion}\label{sec:experiments}
Table~\ref{tab:results} summarizes the results on Okapi dataset and the results from prior work on SCAN dataset~\citep{Keysers2020MeasuringCG, Furrer2020CompositionalGI}. The models' performance is evaluated based on exact match accuracy on test sets averaged over 5 runs. In Okapi dataset, API calls are order invariant with respect to parameters. Therefore, we map the output sequence from each model to an order invariant representation of the API call and then measure the exact match accuracy over the set of parameters and their associated properties and values. 
\\\textbf{\emph{Length} Split vs \emph{Program} Split}\quad 
As discussed in section~\ref{sec:nl2api_tasks}, we want to explore the generalization challenges in Okapi dataset under these two settings. For all these general purpose models the \emph{Length} split is more challenging and even yielding zero accuracy in LSTM+Attention model. The average accuracy of the best performing model on \emph{Program} split and \emph{Length} split of Okapi dataset are 78.5\% and 14.35\%, respectively, indicating that it is more challenging to extrapolate to longer and more complex API calls.
\\\textbf{Pre-trained Language Models}\quad
Table~\ref{tab:results} shows that T5-Base outperforms all other models in being able to extrapolate to longer sequences in test time since it has already seen longer sequences in its pre-training stage. However, given the small size of Okapi dataset, we were not able to fully leverage this generalization property of T5-Base. 

Moreover, we analyzed the output length distribution in all the models and observed that T5-Base tends to generate sequences that are much longer than the gold API calls while Transformer+Copy generates sequences that are exactly in the same length range as the train set. We sampled 100 examples from the erroneous predictions of T5-Base in each Okapi domain for \emph{Length} split. Overall, 21\% of the errors were due to incorrect parameters or incorrect values associated with parameters, 30\% due to missing a parameter from API call, 9\% had an extra parameter in addition to the correct set of parameters, 37\% due to invalid API call or having a long noisy text at the tail, and 3\% due to annotation mistakes. We believe that T5-Base is making a large amount of mistakes in selecting API parameters and their values due to lack of a copying mechanism. 

\begin{figure*}[h]
    \begin{minipage}[b]{0.3\linewidth}
    \centering
    \includegraphics[width=\textwidth]{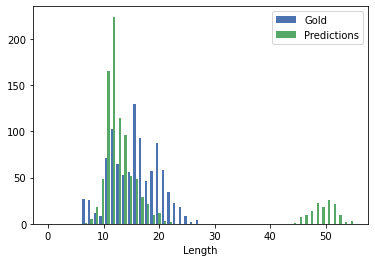}
    \caption*{\small Document}
    \end{minipage}
    \hspace{0.5cm}
    \begin{minipage}[b]{0.3\linewidth}
    \centering
    \includegraphics[width=\textwidth]{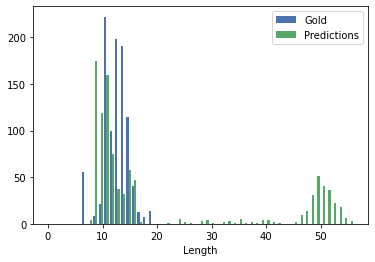}
    \caption*{\small Email}
    \end{minipage}
        \hspace{0.5cm}
    \begin{minipage}[b]{0.3\linewidth}
    \centering
    \includegraphics[width=\textwidth]{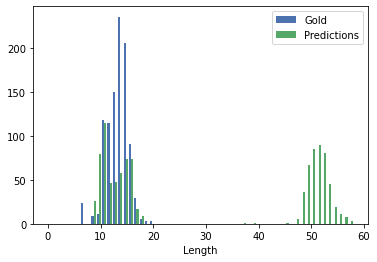}
    \caption*{\small Calendar}
    \end{minipage}
    
    \caption{\small The length distribution of erroneous predictions versus target API calls in \emph{Length} split. The predictions were obtained from fin-tuned T5-Base model .}
    \label{fig:t5-vertical-error-len-dist}
\end{figure*}

\begin{figure*}[h]
    \begin{minipage}[b]{0.3\linewidth}
    \centering
    \includegraphics[width=\textwidth]{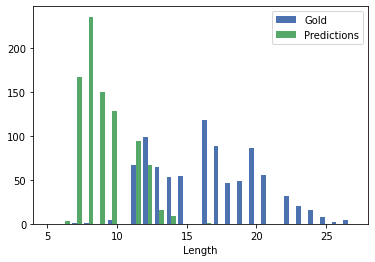}
    \caption*{\small Document}
    \end{minipage}
    \hspace{0.5cm}
    \begin{minipage}[b]{0.3\linewidth}
    \centering
    \includegraphics[width=\textwidth]{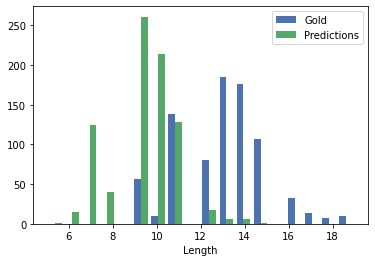}
    \caption*{\small Email}
    \end{minipage}
        \hspace{0.5cm}
    \begin{minipage}[b]{0.3\linewidth}
    \centering
    \includegraphics[width=\textwidth]{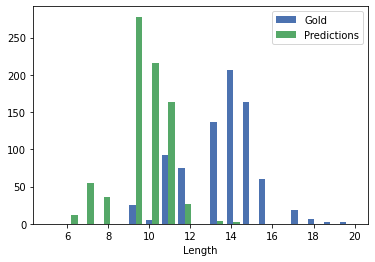}
    \caption*{\small Calendar}
    \end{minipage}
    
    \caption{\small The length distribution of erroneous predictions versus target API calls in \emph{Length} split. The predictions were obtained from Transformer+Copy model.}
    \label{fig:xsp-vertical-error-len-dist}
\end{figure*}

In addition, we sampled 100 examples from the erroneous predictions of T5-Base in \emph{Program} split of each Okapi domain. Similar to the \emph{Length} split we observed a large number of predictions with noisy text at the tail of the API calls (45\%).  Also, 30\% of the errors were due to wrong values as well as spelling mistakes in API parameter names or their values, 10\% due to missing parameters from the API call, 13\% due to having an extra parameter in addition to correct API call, and 2\% due to annotation mistakes. We can see in Figures \ref{fig:t5-vertical-error-len-dist} and \ref{fig:xsp-vertical-error-len-dist} that T5-Base tends to generate sequences that are much longer than the gold API calls while Transformer+Copy generates sequences that are exactly in the same length range as the training set. 
\\
\textbf{Length Extrapolation}\quad Length extrapolation is a difficult task that has not been heavily investigated by the NLP community and is correlated with our goal of expanding to more complex API calls via including more API parameters. Recently, \citealp{dubois-etal-2020-location} and \citealp{newman-etal-2020-eos} investigated the effect of predicting the special \textit{EOS} token in generative models. They claim that training to predict this token causes current models to track generated sequence length in a manner that does not extrapolate to longer sequences. In addition, they showed that the transformer attention has some extrapolation capacity, hence the reason for better performance of Transformer+Copy and T5-Base models on \emph{Length} split. These models have observed API calls up to 14 words and 3 parameters in training set and the set of correct predictions is consisted of API calls with length of 9 to 15 words and mainly 4 parameters. This indicates their limited capacity for extrapolation. While these models fail in \emph{Length} splits for both SCAN and Okapi datasets, several models with specialized architectures~\citep{Shaw2020CompositionalGA, liu2020compositional} have been proposed for SCAN dataset which achieve 100\% accuracy on \emph{Length} and MCD splits. These approaches generally require  manual task specific engineering and are limited by their expressiveness and thus have a low coverage on other datasets. Therefore, we need to propose a solution for length extrapolation in neural generative models or a more specialized model for Okapi dataset to achieve a good performance on \emph{Length} split.
\\
\textbf{Compound Divergence}\quad We apply the distribution based compositionality assessment (DBCA) method~\citep{Keysers2020MeasuringCG} to our \emph{Length} split. This metric claims to provide a quantitative measure that represents the extent of compositional generalization in a dataset split. In Okapi dataset, atoms are API parameters, their properties, and operators such as "\texttt{=,  >, <}" and compounds are the combinations of these atoms. The atom and compound divergence in the Okapi and SCAN (from~\citep{Keysers2020MeasuringCG}) datasets are presented in Table~\ref{tab:dbca_vetical}. Note that the compound divergence in the \emph{Length} split of Okapi datasets is low. Given that we do not observe a  negative correlation between the compound divergence and the exact match accuracy across domains, we conclude that DBCA could be domain-dependent. In addition, we attempted to generate a Maximum Compound Distribution (MCD) split according to \citep{Keysers2020MeasuringCG}, however we were not able to achieve a compound divergence score noticeably higher than the numbers reported in Table~\ref{tab:dbca_vetical}. 

\begin{table}[ht]
\centering
\begin{tabular}{@{}lcc@{}} 
\toprule
\centering
{\small Divergence} & {\small \textbf{Atom} } & {\small \textbf{Compound}}\\ 
\midrule
{\small Document}  & {\small 0.06} & {\small 0.35}   \\
{\small Email} & {\small 0.05} & {\small 0.13}  \\
{\small Calendar}  & {\small 0.04} & {\small 0.13} \\
{\small SCAN} & {\small 0.03} & {\small 0.437}\\
\bottomrule
\end{tabular}
\caption{\label{tab:dbca_vetical}\small DBCA metrics in \emph{Length} split. }
\end{table}

\begin{figure}[h]
    \centering
    \includegraphics[width=0.76\columnwidth]{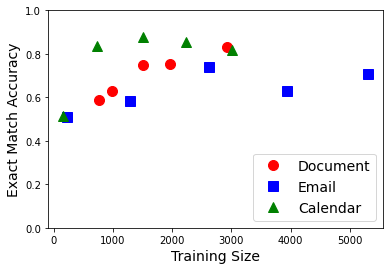}
    \caption{\small Effect of training size on the performance of Transformer+Copy model in \emph{Program} split}
    \label{fig:low_resource_horizontal}
\end{figure}

\textbf{Training Size}\quad We observed very good performance in \emph{Program} split using Transformer+Copy model. In order to analyze the ability of these models to generalize in a low resource setting, we reduced the size of training and validation sets while keeping the exact template distribution as in the full-size dataset. Figure~\ref{fig:low_resource_horizontal} shows that on average the models are able to achieve a similar performance as the full-size dataset setting with only half of that data. 



\section{Conclusion}\label{sec:conclusion}
In this paper we introduce Okapi, a complex and multi-domain natural language to Web API dataset. The dataset is intended to benefit the research community  and enables further studies aiming to solve the compositional challenges in semantic parsing and more specifically, NL2APIs. We defined two challenging and realistic semantic parsing tasks including \emph{Length} split with respect to number of parameters in API calls and \emph{Program} split with respect to API templates. Experimental results on several models that are competitive with publicly available systems for semantic parsing tasks suggest plenty of room for improvement. Even though the \emph{Program} split has a reasonable performance, it is far from perfect specially in its low-resource setting. In this paper, we provide a new resource for studying a different aspect of compositional generalization along the existing datasets. Moreover, we are interested in expanding Okapi to other, especially less-represented, languages in future work to facilitate internationalization efforts.

\bibliographystyle{ACM-Reference-Format}
\bibliography{nl2api}

\appendix

\end{document}